\documentclass[letterpaper, 10 pt, conference]{ieeeconf}  % Comment this line out if you need a4paper

\IEEEoverridecommandlockouts                              % This command is only needed if 
\usepackage{array}    % For better column alignment
\usepackage{multicol} % For spanning text across columns
\usepackage{comment}
\usepackage[table]{xcolor} % important: [table]
\usepackage{tikz}
\usepackage{pgfplots}
\usepackage{graphics} % for pdf, bitmapped graphics files
\usepackage{epsfig} % for postscript graphics files
\usepackage{mathptmx} % assumes new font selection scheme installed
\usepackage{times} % assumes new font selection scheme installed
\usepackage{amsmath} % assumes amsmath package installed
\usepackage{amssymb}  % assumes amsmath package installed
\usepackage{pgfplots}
\pgfplotsset{compat=1.18} % optional but recommended
\definecolor{color2D}{RGB}{176, 196, 206}      % Muted Blue/Grey
\definecolor{colorStereo}{RGB}{250, 218, 193}  % Apricot/Light Orange
\definecolor{colorVR}{RGB}{205, 235, 197}      % Light Green
\let\labelindent\relax
\usepackage{enumitem}
\usepackage{svg}
\usepackage{tablefootnote}
\usepackage{subcaption}
\usepackage{multirow}
\usepackage{colortbl}
\usepackage{siunitx}
\usepackage{booktabs}
\usetikzlibrary{positioning, fit, backgrounds, arrows.meta, shapes.geometric, calc}

\title{\LARGE \bf
VR-DAgger: Immersive VR for Dexterous Data Collection and Uncertainty-Guided On-Policy Correction
}

\author{
René Zurbrügg$^{*}$, Tifanny Portela$^{*}$, Arjun Bhardwaj,\\ Aravind Elanjimattathil Vijayan, Maximum Wilder-Smith, Marco Hutter% <-this % stops a space
\thanks{The Authors are with the Robotics Systems Lab ETH Zürich. 
This research was primarily supported by the ETH AI Center. The project received funding from the ETH Zurich Research Grant No.~22-2 and was supported by ANYbotics AG, the ETH Augmented Reality Research Lab (ETHAR) and Swiss Federal Railways (SBB) via ETH Mobility Initiative. Additional support was provided by NVIDIA, including access to NVIDIA A6000 GPUs. *These authors contributed equally to this work.
}%
}

\begin{document}

\maketitle

% TODO: REMOVE BEFORE SUBMISSION
\thispagestyle{plain}
\pagestyle{plain}

\begingroup
\let\clearpage\relax

%%%%%%%%%%%%%%%%%%%%%%%%%%%%%%%%%%%%%%%%%%%%%%%%%%%%%%%%%%%%%%%%%%%%%%%%%%%%%%%%
% \input{chapters/0_abstract}
% \input{chapters/1_introduction}
% \input{chapters/2_related_work}
% \input{chapters/3_methods}
% \input{chapters/4_results}
% \input{chapters/5_conclusions}
\begin{abstract}
Learning from demonstrations is effective for robotic manipulation, but collecting sufficient task-specific data remains a major bottleneck. Under distribution shift, small errors compound, performance degrades, and expert time is often spent on redundant, low-value corrections instead of the few critical failure cases.

We present \textsc{VR-DAgger}, a human-in-the-loop framework centered on an immersive VR
application for: dexterous teleoperation, demonstration collection, and selective policy
correction. The VR client provides intuitive hand control with synchronized scene
visualization, while a backend workstation runs simulation and learning, enabling
autonomous rollouts without continuous operator oversight. We use Monte Carlo (MC) dropout to score uncertainty during Isaac Lab rollouts of a diffusion policy and select informative failure segments for correction. 
These segments are replayed in VR as clips, where the operator selectively labels
and corrects the policy’s behavior, concentrating supervision where uncertainty is
highest without full-rollout monitoring or a separate intervention classifier.

We evaluate on three dexterous manipulation tasks (Pan pick-and-place, Drawer opening, Valve turning) with a 10-DoF
XHand under standard and challenging initial configurations. Active labeling
consistently improves over behavioral cloning across all tasks, with gains of
up to 23 percentage points.  Compared to unguided human-in-the-loop inspection, \textsc{VR-DAgger}
reduces per-sample collection time by approximately 40\% by focusing review on selected
segments rather than full rollouts. We will release the full framework as open source to
support the community in building on this work.
\end{abstract}

\section{Introduction}
Imitation learning is a practical approach for robotic manipulation, allowing policies to
be learned from demonstrations without dense reward engineering. Recent large behavior
models such as RT-X~\cite{10611477} and Octo~\cite{team2024octo} leverage massive multi-task
datasets, yet reliable performance still demands substantial \emph{task-specific} data, and
collecting this data remains a dominant bottleneck.

This bottleneck has two compounding components. First, real-world demonstration collection
is slow and operationally heavy: scene setup, object placement, and post-failure resets
often dominate wall-clock time \cite{johns2022back}, while safety constraints,
operator oversight, and wear-and-tear make rapid trial-and-error expensive across robotic
platforms. Second, simply collecting more demonstrations exhibits diminishing returns.
Behavioral cloning performs best near the demonstrated state distribution, but small
prediction errors compound over time, pushing the robot into underrepresented regions
where failures become likely~\cite{dagger}. Addressing only one component is insufficient:
accelerating collection without targeting informative states wastes human effort, while
smarter selection without an efficient collection medium leaves the human bottleneck
intact.

\begin{figure}[t!]
    
    \centering
    \includegraphics[width=1.0\linewidth]{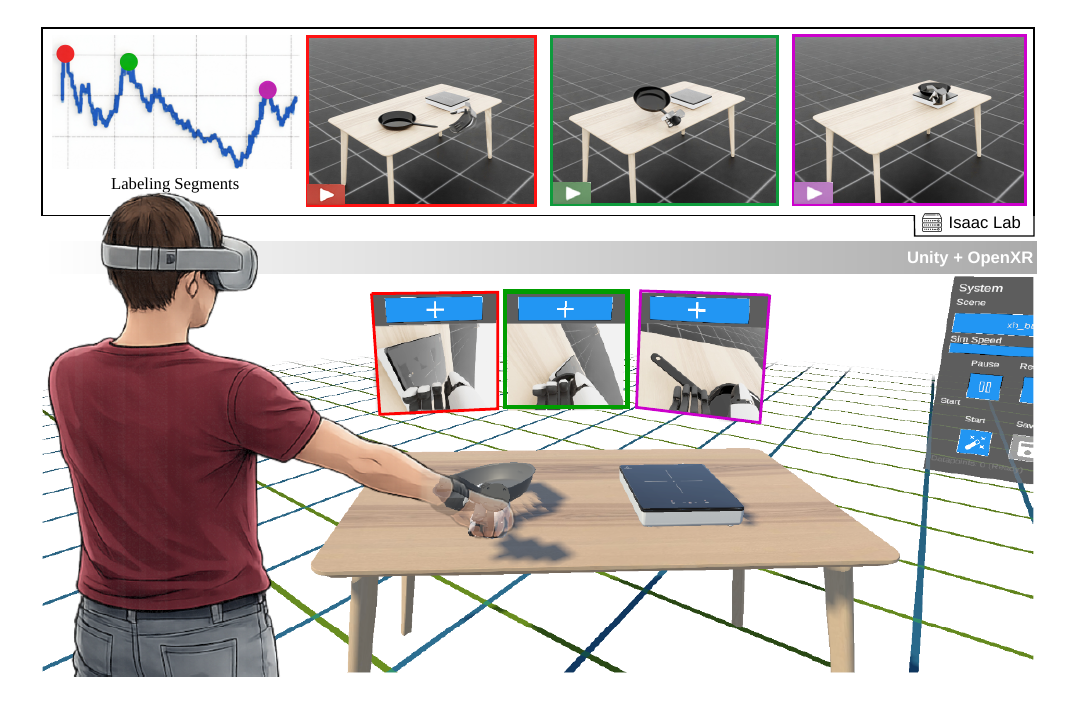}
        \caption{\textbf{\textsc{VR-DAgger}: portable, uncertainty-guided VR data collection.} A task-agnostic VR application integrates directly with any Isaac Lab scene, providing immersive dexterous teleoperation. The policy rolls out autonomously; MC dropout uncertainty selects the most informative failure segments as short snippets for the operator to review and correct in VR, without full-rollout monitoring.}

    % \caption{\textbf{Human-in-the-loop VR data collection with active uncertainty-guided
    % interventions.} The policy rolls out in simulation; Monte Carlo (MC) dropout uncertainty surfaces the
    % most informative segments as short snippets for the operator to review and correct in VR. }
    \label{fig:teaser}
    \vspace{-0.6cm}
\end{figure}
Simulation alleviates the first component: resets are instantaneous, hardware risk is removed, and rollouts can run unattended. However, simulation alone does not resolve the second component: \emph{which} interactions should a person correct, and \emph{when} should they intervene.

Existing approaches only partially address this trade-off. Many VR-based data collection
pipelines emphasize offline demonstration collection~\cite{park2024dexhub}, which can
still consume substantial operator time on redundant or low-information states. On-policy
human-in-the-loop methods reduce compounding error by collecting corrections on the
learner's rollouts, but they often either (i) require synchronous human oversight during
execution~\cite{dagger}, or (ii) rely on a separately trained intervention signal, such as
a failure or risk classifier~\cite{hoque2021thriftydagger}, which adds task-specific
training overhead.
% Most VR-based data collection pipelines focus on offline demonstrations~\cite{park2024dexhub}, which can spend substantial operator time on redundant, low-information states. On-policy human-in-the-loop methods address compounding error by collecting corrections on the learner's rollouts, but they either (i) couple supervision tightly to execution, requiring synchronous oversight during rollouts~\cite{dagger}, or (ii) avoid constant supervision by training a separate intervention signal (e.g., failure/risk classifiers)~\cite{hoque2021thriftydagger}, which introduces additional task-specific data requirements and training cost.

We therefore seek a supervision loop that decouples \emph{execution} from \emph{human
attention}: the policy should explore autonomously, including failures, and the operator
should only be involved in short, high-value correction episodes. For dexterous
manipulation, immersive VR is a natural interface for this setting, as tracked hand motion
can be mapped directly to high-DoF robot hands while preserving intuitive spatial control.
% , so a policy explores autonomously (including failures) and the operator is
% only called in for short, high-value moments using a immersive correction interface.
% For dexterous hand control in particular, immersive VR is a natural fit: it maps tracked
% hand motion directly to high-DoF robot hands while supporting stereo perception and
% vibrotactile feedback~\cite{ding2024bunnyvisionprorealtimebimanualdexterous,
% chen2024arcapcollectinghighqualityhuman}.

% We introduce \textsc{VR-DAgger} (Fig.~\ref{fig:teaser}), a framework designed to mitigate covariate shift in imitation learning. During autonomous policy rollouts in Isaac Sim, we utilize MC dropout~\cite{gal2016dropoutbayesianapproximationrepresenting} to generate a lightweight uncertainty estimate used
% to select informative, high-uncertainty segments of failed trajectories.
% These are shown \emph{after} the rollout
% as short VR snippets, where the operator reviews the outcome and records a corrective
% demonstration, targeting precisely the states where the policy is least certain, without
% full-rollout monitoring. The resulting dataset is aggregated with the original baseline demonstrations to iteratively refine the policy. The system is implemented as a standalone application that connects to a host PC running Isaac Lab. Because it is built on the OpenXR framework, the client can run on any VR or AR headset that supports OpenXR hand tracking. Since it does not rely on stereo rendering, virtual content can blend more naturally with the surrounding environment.

We introduce \textsc{VR-DAgger} (Fig.~\ref{fig:teaser}), an immersive VR framework for
dexterous data collection and interactive policy refinement in Isaac Lab. It supports both
direct teleoperation for demonstration collection and replay-based review of autonomous
policy rollouts for targeted correction. During rollouts, we use MC
dropout~\cite{gal2016dropoutbayesianapproximationrepresenting} as a lightweight
uncertainty estimate to identify informative segments of failed trajectories. These are
replayed \emph{after} the rollout as short VR snippets, which the operator reviews and
corrects through an additional demonstration. This allows the same interface to serve both
as a general-purpose data-collection tool and as an uncertainty-guided supervision
mechanism that focuses human effort on informative failure cases, without continuous
rollout monitoring or a separately trained intervention model. The system is implemented as
a standalone OpenXR-based client connected to a host machine running Isaac Lab. Unlike
approaches that rely on full stereo rendering of the simulation, our interface does not
require headset-specific stereo visualization, making it lighter-weight and easier to
deploy across XR devices. The resulting demonstrations are aggregated to iteratively
refine the policy.
We evaluate on three dexterous manipulation tasks, \emph{Pan} (pick-and-place),
\emph{Drawer} (drawer opening), and \emph{Valve} (valve rotation), with the 10-DoF
\emph{XHand} under standard and
challenging initial configurations, studying the benefit of uncertainty-guided correction
over unguided human-in-the-loop supervision across multiple rounds of active relabeling.
% Across all settings, corrective segments consistently improve on the behavioral cloning
% baseline while requiring substantially less supervisor time than unguided human-in-the-loop (HIL) collection.

In summary, this work contributes:
\begin{enumerate}
    \item An \emph{immersive VR teleoperation environment} compatible with arbitrary
    Isaac Lab scenes, enabling safe data collection and supporting both demonstration
    collection and interactive supervision.
    \item An \emph{active uncertainty-guided labeling method} that uses MC dropout
    uncertainty from the policy itself to surface high-uncertainty snippets for targeted
    human correction, avoiding full-rollout monitoring and eliminating the need to train a
    separate intervention classifier.
    \item \emph{Experimental validation} on three dexterous manipulation tasks
    demonstrating that uncertainty-guided correction improves
    success rate by up to \emph{23 percentage points} over behavioral cloning
   while requiring approximately \emph{40\,\%}
    less collection time per sample than unguided human-in-the-loop supervision.
\end{enumerate}
We will release the framework publicly to facilitate adoption and further research.
\vspace{-0.2cm}

\section{Related Work}

\textbf{Imitation Learning with Diffusion Policies.}
Behavioral Cloning (BC) remains a dominant paradigm for robotic manipulation, achieving strong performance when the test-time state distribution matches the demonstrations. However, in long-horizon tasks, compounding errors induce covariate shift, causing performance to degrade~\cite{dagger}. Recent Large Behavior Models (LBMs), such as RT-X~\cite{10611477} and Octo~\cite{team2024octo}, mitigate this issue through massive multi-task datasets that improve generalization. Despite being effective at scale, these models rely largely on \emph{offline} and static datasets, and lack mechanisms to actively acquire corrective supervision in bottleneck or underrepresented states encountered during deployment. This limitation motivates interactive imitation learning methods~\cite{dagger, kelly2019hg} that adaptively query the expert in states where the policy is most likely to fail, thereby directly addressing the distribution mismatch between expert demonstrations and policy rollouts.

To capture the multimodality of demonstration data and corrective actions provided during interventions, we employ Diffusion Policies~\cite{chi2024diffusionpolicyvisuomotorpolicy}, which represent the policy as a conditional denoising diffusion process. Unlike explicit policies (e.g., Gaussian mixture models) that impose a fixed number of modes and scale poorly to highly multi-modal action distributions, diffusion policies can express complex, continuous action distributions and exhibit high training stability. These models have been shown to outperform the standard BC and LSTM-GMM baselines in robotic manipulation tasks~\cite{chi2024diffusionpolicyvisuomotorpolicy, zhang2024diffusion}.

% \begin{figure*}[th!]
%     \vspace{-1.5cm}
%     \centering
%     \includegraphics[width=0.98\linewidth]{figures/overview.png} % Width set to \textwidth
%     \caption{\textbf{VR-DAgger system architecture and communication stack.} On the server side, the VR-DAgger application runs with Isaac Sim and couples (i) a scene controller for episode management (recording, pause, reset), (ii) a scene-information module that exposes simulator state and camera streams, (iii) GPU-based hand retargeting, and (iv) policy inference (e.g., a diffusion policy) for autonomous rollouts. On the client side, a Unity VR application provides the in-headset user interface and connects to the server via REST for high-level commands/metadata and UDP sockets for low-latency data streaming. The VR device (Meta Quest 3) supplies head and hand tracking through the Oculus runtime. Optionally, a lightweight web UI interfaces with the server through a REST API for configuration and experiment control.}
%     \label{fig:vrdagger_overview}
%     \vspace{-0.5cm}
% \end{figure*}

\begin{figure*}[th!]
    \vspace{0.0pt}
    \centering \hspace{-7mm}
    \includegraphics[width=1.0\linewidth]{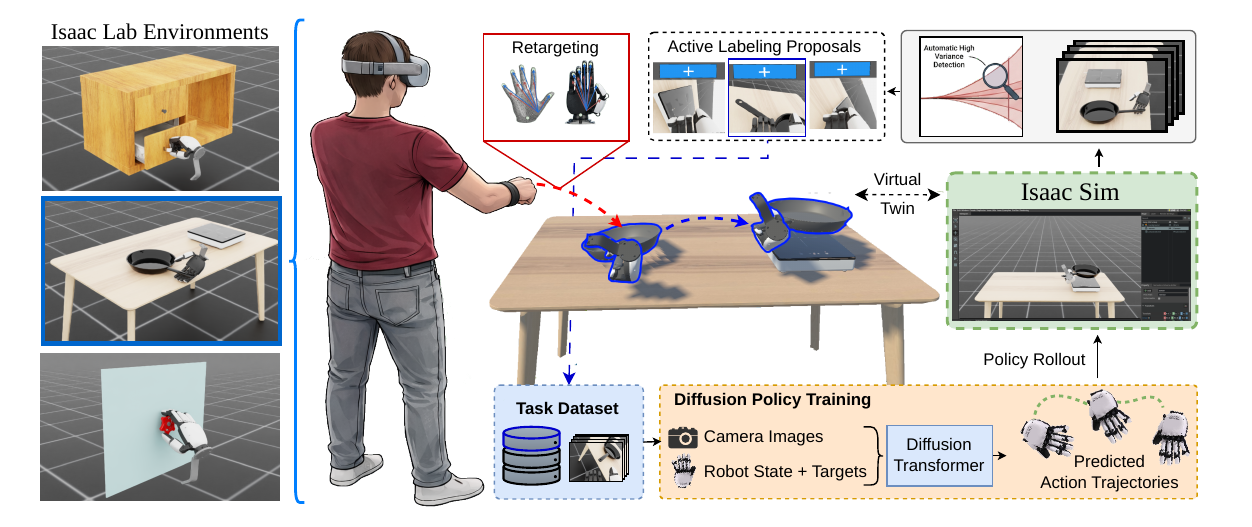} % Width set to \textwidth
    \caption{\textbf{\textsc{VR-DAgger} system overview.} A client-server architecture decouples
    interactive VR teleoperation from simulation and learning. The server runs Isaac Lab,
    policy inference, and MC dropout uncertainty estimation; the Meta Quest client streams
    hand and head tracking for retargeting and visualizes the live scene state. After each
    failed rollout, the highest-uncertainty segment is extracted and presented as a short
    snippet for the operator to review and correct via VR teleoperation. The resulting
    corrective demonstration is appended to the training dataset before the next training
    round.}
    \label{fig:vr_overview}
    \vspace{-0.5cm}
\end{figure*}

\noindent\textbf{Human-in-the-loop Corrections.} Interactive IL reduces supervision cost by querying the expert only in states visited by the learned policy. The seminal DAgger algorithm~\cite{dagger} requires an expert to continuously label states visited by the policy, which places a heavy burden on the human supervisor. To reduce this cost, variants like HG-DAgger (Human-Gated DAgger)~\cite{kelly2019hg} and ThriftyDAgger~\cite{hoque2021thriftydagger} introduce mechanisms where the human or a risk-estimator decides when to intervene. Shared autonomy approaches further attempt to blend human and robot control, often minimizing interventions to only the most critical states~\cite{lazydagger}.

\noindent\textbf{Automatic Credit Assignment.} Identification of portions of a trajectory that induce task failure is a central challenge in the selective querying of human experts for intervention. This problem is particularly challenging in long-horizon tasks, where manually annotating step-by-step supervision is impractical. Prior work such as EnsembleDAgger~\cite{menda2019ensembledagger} approximates a Gaussian Process using an ensemble of neural networks and leverages predictive variance as a confidence metric to determine when to trust the learned policy. To facilitate deployment in safety-critical systems, expert controllers based on robust control barrier functions (CBFs) have been proposed to enforce safety constraints during IL, providing formal safety guarantees for IL policies. However, such methods are ill-suited for tasks where safety constraints are difficult to specify analytically and failure modes are task-dependent.

\noindent\textbf{Uncertainty-Guided Credit Assignment.}
We address credit assignment through an uncertainty-guided supervision loop based on Monte Carlo dropout~\cite{gal2016dropoutbayesianapproximationrepresenting}. During autonomous rollouts, the learner identifies high-uncertainty trajectory segments and extracts them as short demonstration snippets for human correction. This decouples execution from supervision: the policy can explore asynchronously, while operator attention is reserved for the few bottleneck states where it fails or becomes uncertain.

Unlike approaches that mitigate the data demands of imitation learning by collecting a single demonstration and algorithmically expanding it into a synthetic dataset~\cite{yang2024arcadescalabledemonstrationcollection, george2023actplaysingledemonstration}, \textsc{VR-DAgger} targets distribution shift by requesting corrective input precisely at the policy’s failure modes. Systems such as DexHub~\cite{park2024dexhub} show that large-scale simulated datasets collected via immersive interfaces can yield robust policies, but these approaches primarily emphasize \emph{offline} data collection and open-loop training. They do not address the \emph{online} corrective loop in which a deployed policy is monitored and selectively intervened upon during execution. In contrast, \textsc{VR-DAgger} closes the loop by using uncertainty to decide \emph{when} to request supervision and \emph{which} moments to label. We posit that this targeted intervention paradigm yields higher-quality corrective demonstrations while improving data efficiency during policy refinement.

\noindent\textbf{VR/AR Teleoperation.}
VR and AR teleoperation systems provide an intuitive interface and enhance the spatial awareness required for rapid corrective interventions, making them well suited for delivering the short corrective snippets required by our supervision loop. Recent advances in VR and AR teleoperation for robot learning can be broadly categorized into two paradigms: direct control of physical robots and manipulation of simulated digital twins. The former leverages immersive headsets to directly stream human movements to real-world robots in real-time, allowing operators to perceive the environment and gather demonstrations directly on the target hardware~\cite{zhang2018deepimitationlearningcomplex, arunachalam2022holodexteachingdexterityimmersive, ding2024bunnyvisionprorealtimebimanualdexterous}. In the latter paradigm, systems overlay a virtual robot or digital twin onto the real world environment~\cite{duan2023ar2d2trainingrobotrobot, nechyporenko2024armadaaugmentedrealityrobot} or utilize fully simulated virtual spaces~\cite{george2023actplaysingledemonstration}. This allows human operators to control the simulated asset with their hands or AR/VR controllers, capture high-quality kinematic data and generate demonstration datasets without requiring access to a physical robot~\cite{duan2023ar2d2trainingrobotrobot, nechyporenko2024armadaaugmentedrealityrobot, george2023actplaysingledemonstration}. To ensure that the collected data are physically executable, ARCap~\cite{chen2024arcapcollectinghighqualityhuman} actively predicts physical collisions and kinematic constraints, augmenting user perception through visual cues in the headset and controller-based haptic feedback during the manipulation of a simulated digital twin. Similarly, to enhance interaction awareness during the direct teleoperation of real robots, Bunny-VisionPro~\cite{ding2024bunnyvisionprorealtimebimanualdexterous} integrates finger cots with vibration actuators to provide the operator with localized tactile feedback. While prior systems focus on teleoperation and demonstration capture, our work repurposes immersive interfaces as a supervisory layer to provide targeted corrective demonstrations during policy evaluation.

\section{Method}
\label{sec:method}

\textsc{VR-DAgger} is a human-in-the-loop imitation learning framework for
manipulation that combines immersive VR teleoperation with uncertainty-guided active
labeling. The system follows an iterative loop: a visuomotor policy is trained on
collected demonstrations, rolled out in simulation, and on failure, the most
uncertain segments are identified and shown to a supervisor for correction.
The resulting corrective demonstrations are added to the dataset and training continues.
Fig.~\ref{fig:vr_overview} gives an overview of the full pipeline.

\subsection{VR System Overview}
\textsc{VR-DAgger} adopts a client-server architecture that decouples interactive VR
teleoperation from compute-intensive simulation and learning. The client is an OpenXR
application built on top of~\cite{max_vr} and running on a Meta Quest headset, providing
the user interface, hand/head tracking, and a synchronized visualization of the task
environment. Unlike existing VR teleoperation systems which stream only stereo rendering~\cite{mittal2025isaac} or portions of the scene~\cite{zhang2018deepimitationlearningcomplex}, the standalone client can ingest and synchronize \emph{any} USD scene. This lightens the load on both the network and the server PC, freeing up more resources for training. Instead the imported USD scenes are linked to the simulated environment via lightweight object and robot poses As a result, the VR front-end is not tied to a specific
backend and is compatible with other simulators, provided they (i) accept the required
action interface (e.g., end-effector deltas and/or joint targets) and (ii) stream the
poses of the robot and relevant scene objects. The system likewise supports a variety of
end-effector morphologies, from dexterous multi-finger hands to parallel-jaw grippers
(Fig.~\ref{fig:hands}).

In our implementation, the server runs Isaac Lab~\cite{mittal2025isaac} together with
retargeting, policy inference, logging, and (asynchronous) training and relabeling jobs.
For real-time state exchange, we use ROS~1/UDP as the primary data plane, streaming tracked
poses to the server and returning robot/scene state for visualization; we also support
ROS~2 and direct UDP streaming. A REST interface serves as the control plane for simulation
commands (e.g., reset/pause/record), episode summaries, and training dashboards. This
separation keeps the headset loop responsive while enabling headless, parallel simulation
rollouts and server-side diffusion policy training.

\begin{figure}[t!]
    \centering
    \includegraphics[width=0.9\linewidth]{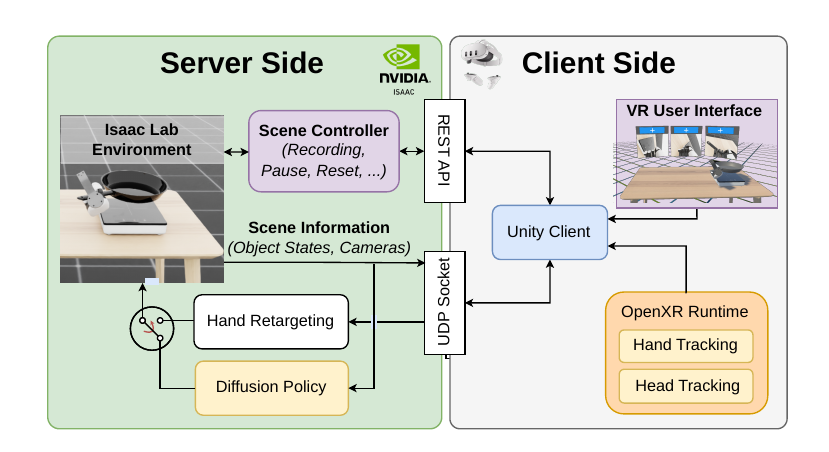}
    \caption{\textbf{VR client interface.} The server runs the Isaac Lab environment,
    scene control (recording, pause, reset), hand retargeting, and diffusion
    policy inference. The client is a native Unity VR application on a Meta Quest headset that
    receives scene state via UDP and sends hand and head tracking from the OpenXR runtime.
    The system is compatible with any Isaac Lab environment consisting of
    rigid bodies, as the scene USD is used to synchronize the environment.}
    \label{fig:vr_client}
    \vspace{-5mm}
\end{figure}
\begin{figure}[b!]
    \centering
    \includegraphics[width=1.0\linewidth]{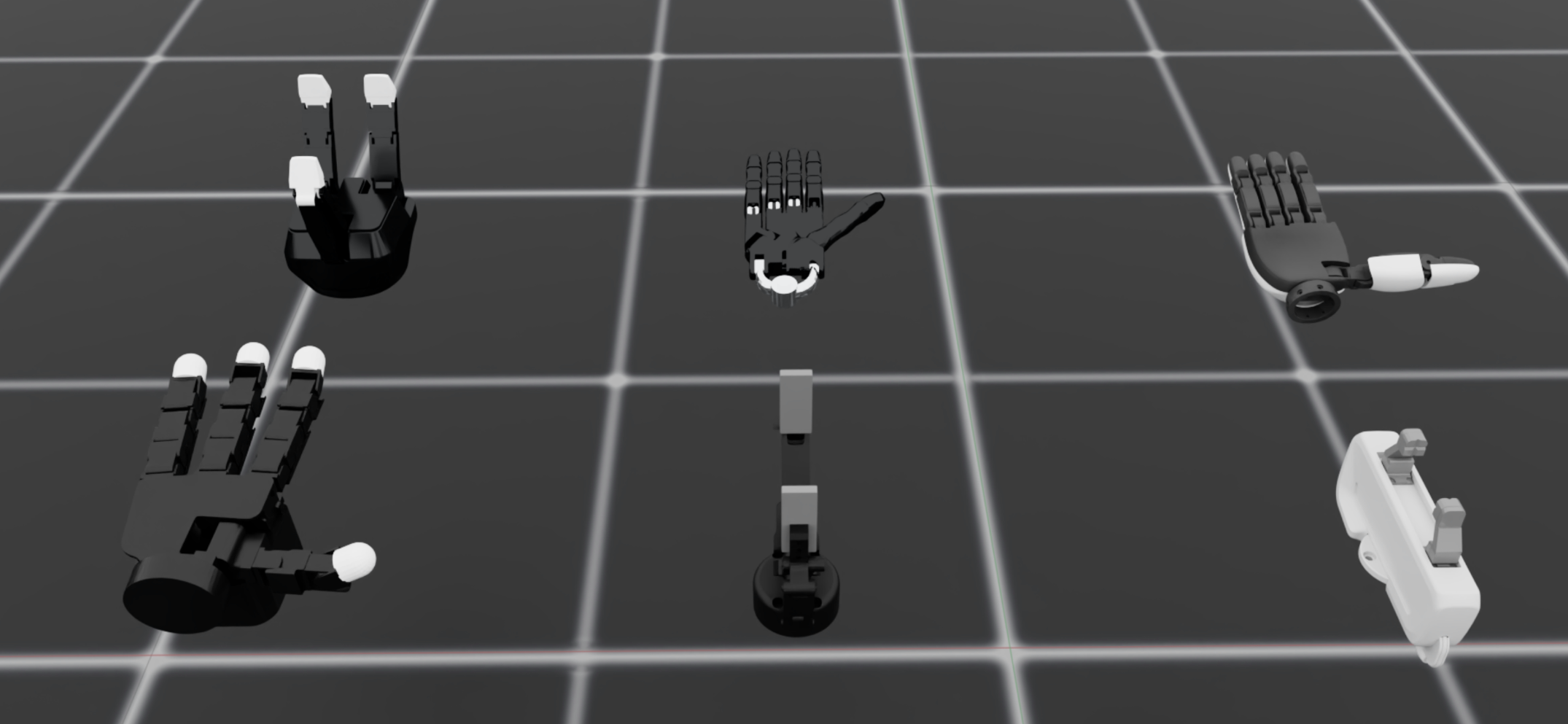}
    \caption{\textbf{Supported end-effectors.} Our VR teleoperation app and 
retargeting pipeline support a variety of gripper morphologies, enabling 
data collection and policy learning across different robot 
configurations.}
    \label{fig:hands}
    % \vspace{-5mm}
\end{figure}

\subsection{Teleoperation and Data Collection}
The operator interacts with the system in three modes: (i) \emph{teleoperation}, where
tracked hand motion is retargeted to robot commands; (ii) \emph{intervention}, where the
policy executes actions and the operator temporarily overrides control at critical failure
points; and (iii) \emph{active labeling}, where the system queries the operator for
targeted annotations or corrective segments selected from policy rollouts (e.g., based on
uncertainty or failure cues), without requiring the operator to supervise entire episodes
synchronously.

In all modes, the backend logs time-aligned observations and executed actions at each
timestep, including rendered RGB views (wrist-mounted and external cameras), robot
proprioception (joint positions), commanded end-effector deltas and/or joint targets, and
episode metadata (mode, timestamps, and labeling state) for aggregation and training.

To reduce invalid demonstrations and allow for easy human takeover, we continuously monitor the alignment between the
tracked human hand and the robot end-effector pose used for retargeting. The headset
displays a green/red overlay and we pause recording when the mean keypoint error exceeds a
threshold $\tau$ (15\,cm in all experiments).

Finally, we optionally provide vibrotactile feedback through a wearable glove~\cite{bhaptics2026} by mapping simulated
contact signals (e.g., per-fingertip contact force magnitude) to actuator amplitudes,
supporting precise manipulation during teleoperation and interventions. These tactile signals are supplemented with color indicators on each finger indicating the intensity of the signal.

\noindent\textbf{Motion Retargeting.}\label{sec:retargeting}
Following prior work~\cite{handa2020dexpilot, qin2021dexmv, qin2023anyteleop}, we formulate
retargeting as an optimization problem that maps tracked human hand motion to the robot
hand despite differing kinematics and joint limits. Previous approaches decouple finger
retargeting from wrist tracking, typically using the wrist pose estimate from the hand pose
estimator directly. In contrast, we jointly optimize the robot joint positions
$\mathbf{q}_t$ and a global wrist pose $\mathbf{T}_t=(\mathbf{R}_t,\mathbf{t}_t)\in SE(3)$.
\begin{figure}[t!]
    \centering
    \includegraphics[width=0.5\linewidth]{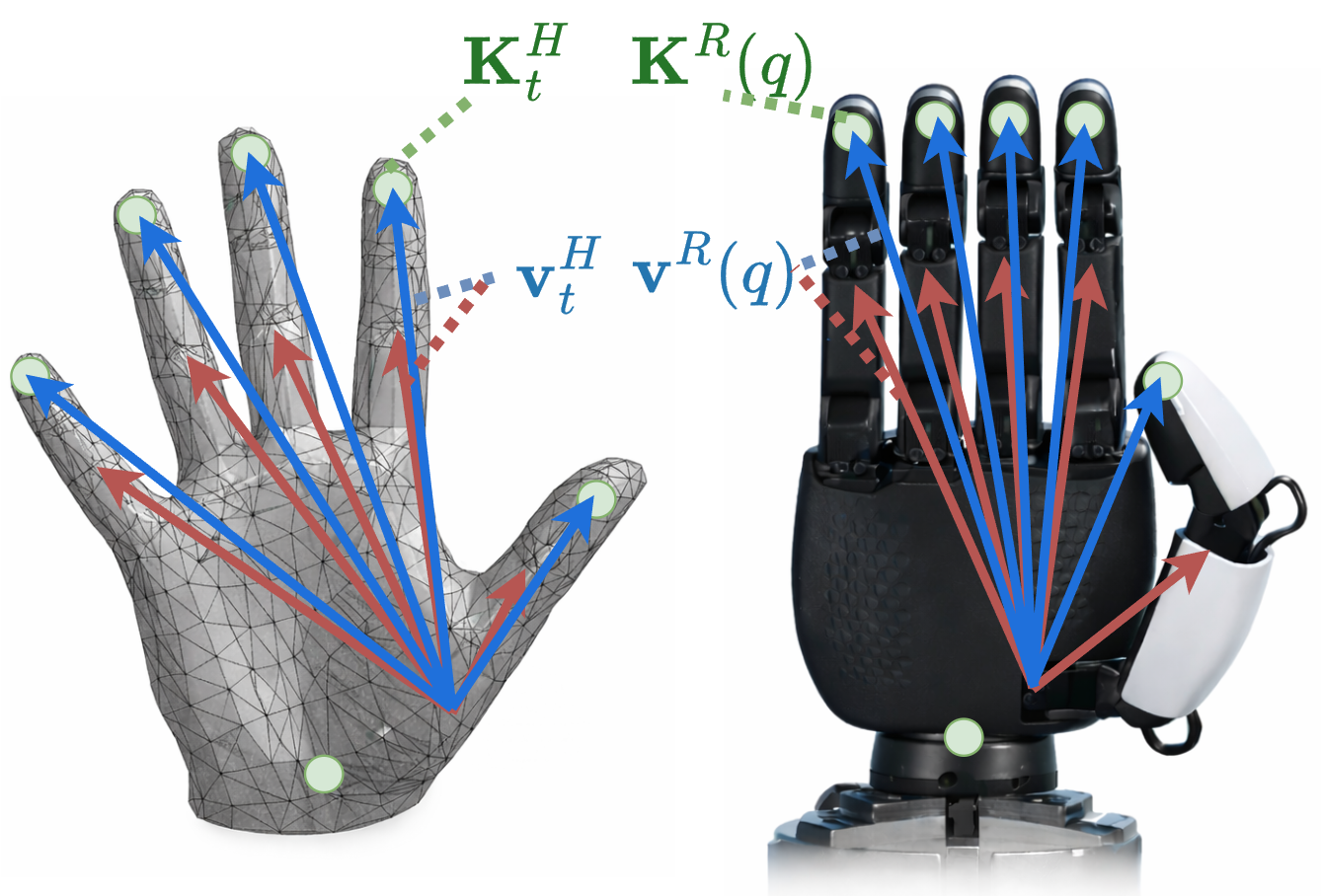}
    \caption{\textbf{Real-time GPU-based retargeting pipeline.} Tracked human keypoints
    $\color[rgb]{0.1216,0.4667,0.1059}\mathbf{K}^H_t$ and task-space key vectors
    $\color[rgb]{0.1216,0.4667,0.7059}\mathbf{v}^H_t$ are mapped to their robot counterparts
    $\color[rgb]{0.1216,0.4667,0.1059}\mathbf{K}^R(\mathbf{q})$ and
    $\color[rgb]{0.1216,0.4667,0.7059}\mathbf{v}^R(\mathbf{q})$, computed via forward
    kinematics. Desired joint targets $\mathbf{q}^{\mathrm{des}}_t$ are obtained by
    solving a weighted least-squares objective with joint-limit constraints
    (Eq.~\eqref{eq:retarget}).}
    \label{fig:retargeting}
    \vspace{-7mm}
\end{figure}

\begin{figure*}[t!]
    \centering
    \includegraphics[width=1.0\linewidth]{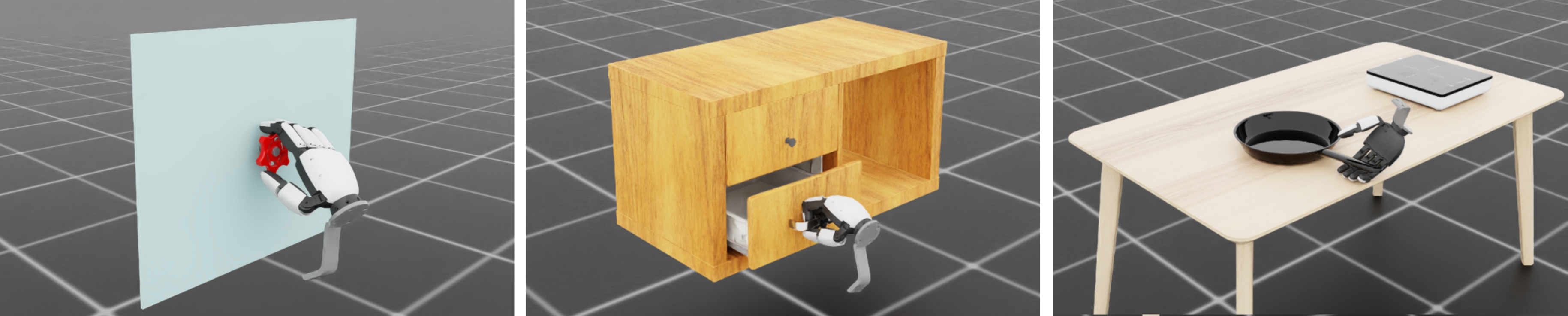}
 \caption{\textbf{Evaluation Environments.} Overview of the three manipulation tasks used for policy evaluation. \emph{(a) Pan:} A long-horizon pick-and-place task in which a robotic arm must grasp a pan from a cluttered table surface and place it onto a target burner, requiring precise grasping and object re-positioning. \emph{(b) Valve:} A revolute joint manipulation task in which a dexterous robotic hand must rotate a wall-mounted valve, testing the ability to apply controlled rotational force to non-prehensile objects. \emph{(c) Drawer:} A task in which a dexterous hand must identify and pull open a drawer on a multi-drawer cabinet, evaluating generalization to prismatic joint manipulation and contact-rich interactions.}
    \label{fig:tasks_overview}
    
    \vspace{-5mm}
\end{figure*}

We introduce \emph{anchor} keypoints that ground the wrist pose in a shared
reference frame (e.g., the headset/camera or world frame), since relative keypoint vectors
alone are translation-invariant. Let $\mathbf{K}^H_{j,t}\in\mathbb{R}^3$ denote the
3D position of the $j$-th tracked \emph{human} anchor keypoint at time $t$ in this shared
frame, and let $\mathbf{K}^R_j(\mathbf{q}_t)\in\mathbb{R}^3$ denote the corresponding
\emph{robot} anchor keypoint position obtained via forward kinematics, expressed in the
robot wrist frame. The global wrist pose $\mathbf{T}_t$ maps robot-frame anchor points
into the shared frame via $\mathbf{R}_t\,\mathbf{K}^R_j(\mathbf{q}_t)+\mathbf{t}_t$.
At each time step $t$, we solve:
% \begin{align}
% \min_{\mathbf{q}_t,\, \mathbf{T}_t} \quad
% & \underbrace{
% \sum_{i=1}^{N} w_i
% \left\lVert \mathbf{v}^H_{i,t} - \mathbf{R}_t \, \mathbf{v}^R_i(\mathbf{q}_t)
% \right\rVert^2
% }_{\text{finger configuration}} + \lambda_a \underbrace{
% \sum_{j=1}^{M}
% \left\lVert \mathbf{K}^H_{j,t} - \big(\mathbf{R}_t\,\mathbf{K}^R_j(\mathbf{q}_t) + \mathbf{t}_t\big)
% \right\rVert^2
% }_{\text{absolute pose alignment}} \nonumber \\
% & + \lambda_q \left\lVert \mathbf{q}_t - \mathbf{q}_{t-1} \right\rVert^2 
% + \lambda_T \left\lVert \boldsymbol{\xi}_t \right\rVert^2 \nonumber \\
% & \text{s.t.} \quad \mathbf{q}_l \leq \mathbf{q}_t \leq \mathbf{q}_u,
% \label{eq:retarget}
% \end{align}
\begin{align}\hspace{-2mm}
\small
\min_{\mathbf{q}_t, \mathbf{T}_t} \;
& \textstyle\sum_{i=1}^{N} w_i 
\left\lVert \mathbf{v}^H_{i,t} - \mathbf{R}_t \mathbf{v}^R_i(\mathbf{q}_t) \right\rVert^2
+ \lambda_a \textstyle\sum_{j=1}^{M}
\left\lVert \mathbf{K}^H_{j,t} - \mathbf{R}_t \mathbf{K}^R_j(\mathbf{q}_t) - \mathbf{t}_t \right\rVert^2 \nonumber \\
& + \lambda_q \left\lVert \mathbf{q}_t - \mathbf{q}_{t-1} \right\rVert^2 
+ \lambda_T \left\lVert \boldsymbol{\xi}_t \right\rVert^2, \quad
\text{s.t.} \; \mathbf{q}_l \leq \mathbf{q}_t \leq \mathbf{q}_u
\label{eq:retarget}
\end{align}
where $\mathbf{v}^H_{i,t}$ is the $i$-th human relative keypoint vector,
$\mathbf{v}^R_i(\mathbf{q}_t)$ is the corresponding \emph{robot} relative keypoint vector in
the wrist frame, and $\mathbf{q}_l,\mathbf{q}_u$ are joint limits. The weights $w_i$
emphasize task-relevant keypoints (e.g., fingertips). The anchor term aligns $M$ designated
keypoints (e.g., palm center and selected fingertips), coupling wrist pose and finger
retargeting.

We improve temporal smoothness by regularizing joint and pose increments using
$\boldsymbol{\xi}_t=\log\!\left(\mathbf{T}_{t-1}^{-1}\mathbf{T}_t\right)\in\mathbb{R}^6$
(with separate scale factors for translation and rotation components), and parameterize
$\mathbf{R}_t$ with a continuous 6D rotation representation~\cite{zhou2020continuityrotationrepresentationsneural}.
We warm-start from $(\mathbf{q}_{t-1},\mathbf{T}_{t-1})$ and solve with a fixed iteration
budget on GPU using NVIDIA Warp to maintain real-time rates (Fig.~\ref{fig:retargeting}).

\subsection{Policy Learning}
\label{sec:objective}
The imitation learning policy predicts a horizon-$H$ trajectory of hand joint and end-effector pose commands  in the current end-effector frame 
from a short history of observations:
\begin{equation}
\boldsymbol{\tau}_{t:t+H} \sim \pi_\theta(\boldsymbol{\tau}\mid o_{t-n_h:t}),
\label{eq:diffusion}
\end{equation}
where $o_{t-n_h:t}$ contains the last $n_h{=}2$ steps of RGB images and
proprioception (hand joint positions and 6D end-effector pose expressed in the current
end-effector frame), and $H{=}16$. Orientations are represented with a continuous 6D
rotation parametrization~\cite{zhou2020continuityrotationrepresentationsneural}. We adopt
a DDPM-style diffusion objective: at each training step a diffusion timestep is sampled,
Gaussian noise is added to the ground-truth trajectory, and the model is trained to
minimize the MSE between predicted and true noise residual $\epsilon$. We use the
Diffusion Policy framework~\cite{chi2024diffusionpolicyvisuomotorpolicy} throughout.

\subsection{Uncertainty-Guided Active Relabeling}
\label{sec:active_relabeling}

A key limitation of standard human-in-the-loop approaches is that the supervisor must
watch full rollouts to identify where to intervene, which is slow and cognitively
demanding. \textsc{VR-DAgger} replaces this with a principled active relabeling step that
automatically identifies the most uncertain segment of a failed episode and presents it to
the human as a short snippet for correction.

\noindent\textbf{Uncertainty estimation via MC dropout.}
After each rollout, we estimate policy uncertainty at every timestep using Monte Carlo
dropout~\cite{gal2016dropoutbayesianapproximationrepresenting}. We perform $N=10$ stochastic forward passes through the
diffusion policy with dropout enabled at inference time (all layers that use dropout
during training), yielding $N$ trajectory samples
$\{\hat{\boldsymbol{\tau}}^{(i)}_{t:t+H}\}_{i=1}^N$ at each timestep $t$. The uncertainty
score is the mean predictive variance across action dimensions, computed in the same
normalized action space used for training:
\begin{equation}
u_t = \frac{1}{D}\sum_{d=1}^{D} \mathrm{Var}_{i}\!\left[\hat{\tau}^{(i)}_{t,d}\right],
\label{eq:uncertainty}
\end{equation}
where $D$ is the action dimensionality. This gives a scalar uncertainty trace $\{u_t\}$
over the full episode.

\noindent\textbf{Segment selection and human correction.}
Active relabeling is invoked only for failed episodes. For each failed rollout, we compute
an uncertainty score $u_t$ at every timestep and select the three peak timesteps
$t_k^* \in \operatorname{Top}\text{-}3_t\, u_t$. For each $t_k^*$, we extract a short
context window of 2\,s centered at $t_k^*$ and present the resulting
snippets to the supervisor as 2D playback videos inside the VR interface. The supervisor reviews the three candidate snippets and selects
the segment that appears most in need of correction, without watching the full trajectory. Starting from the selected timestep, the supervisor provides a corrective demonstration by
taking over teleoperation control for the remainder of the episode. The resulting
corrective segment is appended to the training dataset, after a pre-defined number of data samples (50)  the policy is
retrained.

\begin{table*}[t]
    \centering
   \caption{\textbf{Policy performance across demonstration configurations, tasks, and difficulty levels.}
    Each row corresponds to a training configuration defined by the number of initial offline demonstrations and corrective (on-policy) segments.
    \emph{Behavioral Cloning} (BC) rows use only offline data; \emph{Corrective} rows use \textsc{VR-DAgger} uncertainty-guided active relabeling; \emph{HIL} rows use unguided human-in-the-loop inspection.
    For each task (\emph{Pan}, \emph{Drawer}, \emph{Valve}) and difficulty (\emph{Default}, \emph{Hard}), we report task success rate (\%), subtask success rate (\%) and respective standard deviations, evaluated after 1200 training epochs across 256 parallel environments for three different seeds.
    Subtask SR measures grasp success for Pan, grasp success for Drawer, and partial valve rotation for Valve.}
    \label{tab:vrdagger_results}
    \resizebox{\textwidth}{!}{%
    \setlength{\tabcolsep}{3pt}
    \begin{tabular}{l|cc||cc|cc||cc|cc||cc|cc}
        \toprule
        \multirow{4}{*}{\textbf{Experiment}}
            & \multicolumn{2}{c|}{\textbf{Demonstrations}}
            & \multicolumn{12}{c}{\textbf{Tasks}} \\
        \cmidrule(lr){2-3} \cmidrule(lr){4-15}
            & \multirow{3}{*}{\textbf{Initial}}
            & \multirow{3}{*}{\textbf{Corrective}}
            & \multicolumn{4}{c|}{\textbf{Pan on Burner}} & \multicolumn{4}{c|}{\textbf{Drawer}} & \multicolumn{4}{c}{\textbf{Valve}} \\
        \cmidrule(lr){4-7} \cmidrule(lr){8-11} \cmidrule(lr){12-15}
            & &
            & \multicolumn{2}{c|}{\textbf{Default}} & \multicolumn{2}{c|}{\textbf{Hard}} & \multicolumn{2}{c|}{\textbf{Default}} & \multicolumn{2}{c|}{\textbf{Hard}} & \multicolumn{2}{c|}{\textbf{Default}} & \multicolumn{2}{c}{\textbf{Hard}} \\
        \cmidrule(lr){4-5} \cmidrule(lr){6-7} \cmidrule(lr){8-9} \cmidrule(lr){10-11} \cmidrule(lr){12-13} \cmidrule(lr){14-15}
            & &
            & \textbf{Task} (\%) & \textbf{Subtask} (\%)
            & \textbf{Task} (\%) & \textbf{Subtask} (\%)
            & \textbf{Task} (\%) & \textbf{Subtask} (\%)
            & \textbf{Task} (\%) & \textbf{Subtask} (\%)
            & \textbf{Task} (\%) & \textbf{Subtask} (\%)
            & \textbf{Task} (\%) & \textbf{Subtask} (\%) \\
        \midrule
        
        % \rowcolor{blue!10}
        \rowcolor{blue!5} 
        BC    & 50   & --     & $49 _{\pm 4.1}$ & $72 _{\pm 4.1}$ & $18 _{\pm 0.9}$ & $38 _{\pm 2.7}$ & $87 _{\pm 1.6}$ & $92 _{\pm 2.3}$ & $53 _{\pm 5.9}$ & $62 _{\pm 4.9}$ & $84 _{\pm 2.5}$ & $88 _{\pm 1.4}$ & $54 _{\pm 2.3}$ & $59 _{\pm 3.1}$ \\
        \midrule
        \midrule
        
         BC    & 100  & --     & $40 _{\pm 0.9}$ & $66 _{\pm 4.7}$ & $18 _{\pm 3.9}$ & $39 _{\pm 5.5}$ & $92 _{\pm 2.1}$ & $95 _{\pm 0.5}$ & $71 _{\pm 2.3}$ & $78 _{\pm 3.2}$ & $92 _{\pm 0.5}$ & $97 _{\pm 1.6}$ & $72 _{\pm 2.7}$ & $81 _{\pm 1.6}$ \\
        \rowcolor{gray!10}BC    & 150  & --    & $50 _{\pm 0.8}$ & $70 _{\pm 3.4}$ & $30 _{\pm 2.0}$ & $64 _{\pm 4.3}$ & $93 _{\pm 3.3}$ & $97 _{\pm 0.5}$ & $79 _{\pm 2.8}$ & $84 _{\pm 2.8}$ & $\underline{98} _{\pm 1.6}$ & $98 _{\pm 2.1}$ & $84 _{\pm 4.1}$ & $92 _{\pm 3.5}$ \\
        \midrule
        HIL-Inspection          & 50  & 50  & $56 _{\pm 3.0}$ & $86 _{\pm 2.1}$ & $26 _{\pm 1.6}$ & $54 _{\pm 1.6}$ & $91 _{\pm 3.6}$ & $96 _{\pm 1.6}$ & $84 _{\pm 4.4}$ & $89 _{\pm 3.3}$ & $\underline{98} _{\pm 1.6}$ & $99 _{\pm 0.9}$ & $83 _{\pm 2.7}$ & $89 _{\pm 2.7}$ \\
       \rowcolor{gray!10} HIL-Inspection          & 50  & 100  & $\underline{67} _{\pm 5.0}$ & $\underline{91} _{\pm 2.7}$ & $\underline{34} _{\pm 3.6}$ & $\mathbf{68} _{\pm 2.1}$ & $\mathbf{97} _{\pm 1.6}$ & $\underline{98} _{\pm 2.1}$ & $\mathbf{92} _{\pm 1.6}$ & $\mathbf{95} _{\pm 0.9}$ & $\mathbf{99} _{\pm 0.8}$ & $\mathbf{100} _{\pm 0.0}$ & $\mathbf{98} _{\pm 1.4}$ & $\mathbf{98} _{\pm 0.5}$ \\
        \midrule
        Corrective   & 50  & 50     & $54 _{\pm 8.7}$ & $76 _{\pm 2.3}$ & $22 _{\pm 2.5}$ & $54 _{\pm 0.8}$ & $\mathbf{97} _{\pm 1.2}$ & $\underline{98} _{\pm 0.8}$ & $\underline{86} _{\pm 2.7}$ & $\underline{92} _{\pm 1.2}$ & $97 _{\pm 1.8}$ & $\mathbf{100} _{\pm 0.5}$ & ${89} _{\pm 1.2}$ & $\underline{95} _{\pm 2.8}$ \\
        \rowcolor{gray!10}Corrective   & 50  & 100     & $\mathbf{72} _{\pm 2.4}$ & $\mathbf{92} _{\pm 3.0}$ & $\mathbf{36} _{\pm 3.7}$ & $\underline{65} _{\pm 1.6}$ & $\mathbf{97} _{\pm 0.5}$ & $\mathbf{99} _{\pm 0.9}$ & $\underline{86} _{\pm 0.6}$ & $\underline{92} _{\pm 1.7}$ & $97 _{\pm 1.6}$ & $\mathbf{100} _{\pm 0.0}$ & $\underline{92} _{\pm 0.9}$ & $\underline{95} _{\pm 0.0}$ \\
 
        \bottomrule
        
    \end{tabular}
    }
\end{table*}

\section{Results}

\subsection{Tasks and Evaluation Environments}
We evaluate \textsc{VR-DAgger} on three manipulation tasks of varying complexity
(Fig.~\ref{fig:tasks_overview}). \emph{Pan} is a long-horizon pick-and-place task in
which a dexterous hand must grasp a pan from a cluttered surface and place it onto a
target burner, requiring precise grasping and object repositioning. \emph{Drawer} requires
a dexterous hand to identify and pull open a drawer on a multi-drawer cabinet, involving
contact-rich interaction with a prismatic joint. \emph{Valve} is a revolute joint
manipulation task in which a dexterous hand must apply controlled rotational force to a
wall-mounted valve. All three tasks are evaluated on the 10-DoF \emph{XHand} under two
difficulty levels: \emph{Default}, with standard initial configurations, and \emph{Hard},
with greater variation in object placement and initial hand pose.

\subsection{Experimental Setup}

% \begin{figure*}[t]
%     \centering
%     \includegraphics[width=1.0\linewidth]{figures/iros/trend.pdf}
%     \caption{\textbf{Success rate as a function of total demonstrations across tasks and difficulty levels.} Each plot shows success rate (\%) for Pan, Drawer, and Valve as the demonstration budget grows from 50 initial demonstrations to 150 (50 initial + up to 100 corrective segments, added in increments of 25). The leftmost marker denotes behavioral cloning trained on 50 demonstrations only. Solid and dashed lines show \textsc{VR-DAgger} performance on Default and Hard configurations respectively, with each additional round of active relabeling. Gains are consistent across all tasks, with particularly large improvements on Hard configurations (e.g., Drawer Hard: $47.4\%\to84.2\%$).}
%     \label{fig:plot_over_time}
% \end{figure*}
\noindent\textbf{Data collection protocol.}
For each task, we collect an initial set of 50 demonstrations via VR teleoperation. We
then run two rounds of uncertainty-guided active relabeling, adding 50 corrective
segments per round, for a maximum total of 150 demonstrations (50 initial + 100
corrective) per configuration. Crucially, all data collection and active relabeling
is performed exclusively under \emph{Default} difficulty; no corrective labels are collected
under \emph{Hard} configurations. \emph{Hard} difficulty results therefore reflect the policy's ability
to generalize beyond the demonstrated distribution to more challenging initial conditions.

\noindent\textbf{Compared configurations.} We evaluate three training configurations that progressively exercise the components of \textsc{VR-DAgger}. \emph{Behavioral Cloning} (BC) trains solely on offline demonstrations collected through our VR teleoperation interface, with no active relabeling; we evaluate at three budget levels (50, 100, and 150 demonstrations) to quantify the effect of simply scaling offline data. Both \emph{Human-in-the-loop (HIL)} and \emph{Corrective} additionally perform iterative on-policy data collection within the \textsc{VR-DAgger} framework, differing only in how corrective segments are selected. \emph{Corrective} uses our proposed uncertainty-guided active relabeling (Sec.~\ref{sec:active_relabeling}), presenting high-uncertainty snippets for targeted correction, while \emph{HIL} relies on unguided manual inspection: the supervisor watches full rollouts of failed episodes and decides where to intervene. 

\noindent\textbf{Metrics.} For each task and difficulty level, we report three metrics in Table~\ref{tab:vrdagger_results}: \emph{Task Success Rate}, the fraction of episodes completed successfully; \emph{Subtask Success Rate}, measuring completion of intermediate milestones (grasp success for Pan, contact success for Drawer, and partial rotation for Valve); and \emph{Task Completion Time}, the mean simulation time to success across successful episodes. All are evaluated after 1200 training epochs across 256 parallel environments. Additionally, we report \emph{Data Collection Time}, the mean wall-clock time per collected trajectory in Fig.~\ref{fig:data_timings}, capturing the practical cost of each supervision strategy.

\noindent\textbf{Training details.} All diffusion policies are trained for 1200 epochs on a single RTX 4090, requiring approximately 14 hours per configuration. An episode is considered successful if the task is completed within an 8-second time limit. For the Pan task, which requires precise placement, we additionally include an external fixed camera to support placement detection.

% \noindent\textbf{Runs and randomness.}
% Unless stated otherwise, results are averaged over three random seeds and we report mean
% and standard deviation; entries without $\pm$ are single-run evaluations. Random seeds
% control simulator initialization and policy training. Active relabeling is triggered only
% on failed episodes; in each round we relabel up to 25 failed episodes, and if fewer
% failures are available we relabel all of them.

\begin{figure*}[t!]
    \centering
    \includegraphics[width=1.0\linewidth]{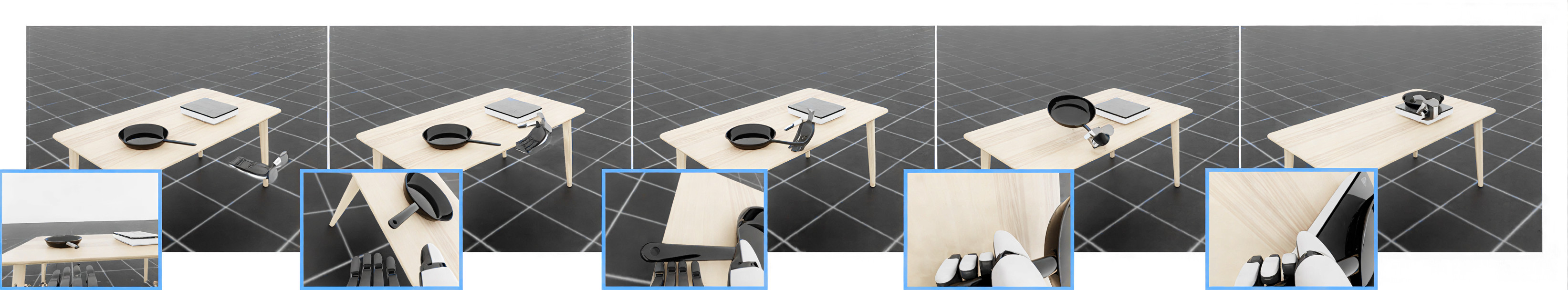}
    \caption{\textbf{Example rollout for the Pan task.} Sequence of global simulator frames (top) with the corresponding wrist-mounted frames (bottom, blue insets) as the robot approaches, grasps, and manipulates the object. During training and inference, the policy integrates both the third-person global perspective for spatial context and the ego-centric wrist views for fine-grained manipulation.}
    \label{fig:pan_rollout}
    \vspace{-5mm}
\end{figure*}

\subsection{Policy Performance}

Table~\ref{tab:vrdagger_results} summarizes success rates across all three tasks and
both difficulty levels for the \emph{Behavioral Cloning} (BC), \emph{HIL}, and \emph{Corrective}
configurations.
Fig.~\ref{fig:performance} provides a visual breakdown including subtask success rates.

\noindent\textbf{Active relabeling consistently improves performance.}
Across all tasks, both Corrective and \textit{HIL} improve success rate over Behavioral Cloning.
The key mechanism is exposure to the \emph{failure regime}: on-policy collection, whether
uncertainty-guided or not, naturally surfaces states where the current policy diverges
from the expert distribution -- states that are underrepresented in offline demonstrations.
Adding corrective demonstrations in these states teaches the policy to recover, closing
the loop that BC leaves open. On Default difficulty, where \textit{BC} already achieves moderate
success (87\% on Drawer, 84\% on Valve, 49\% on Pan with 50 demos), the gains are meaningful but
moderate: Corrective pushes Drawer Default to 97\%, Pan Default to 54\% and Valve Default to 97\%.

The effect is also present on Hard configurations, even though no
corrective data is ever collected under Hard difficulty. Hard results measure
generalization: the policy, having learned to recover from failure under Default
conditions, transfers this recovery behavior to harder initial states it has never been
supervised on. On Valve Hard, \textit{BC} (50 demos) achieves 54\%, while Corrective (50\,+\,50)
reaches 89\% (+35\,pp). On Drawer Hard, \textit{BC} achieves 53\% and Corrective
50\,+\,50 reaches 86\% (+33\,pp). On Pan Hard, the same budget yields a modest lift
(18\%$\to$22\%); doubling the corrective budget to 100 segments raises it to 36\%
(+18\,pp), underscoring that the long-horizon structure of Pan requires more extensive
failure-regime coverage to generalize.

\noindent\textbf{Corrective vs.\ HIL.}
HIL serves as a human upper bound: a supervisor with full rollout visibility and no
budget constraint on attention will naturally cover more failure modes. Accordingly, at
lower label budgets (50\,+\,50), HIL remains competitive. Interestingly, Corrective outperforms HIL on some conditions (Drawer Default: 97\% vs.\ 91\%, Valve Hard: 89\%
vs.\ 83\%, Drawer Hard: 86\% vs.\ 84\%), showing that targeted snippet selection is
sufficient to approach or overcome the quality of full rollout supervision. At
the larger budget (50\,+\,100), HIL achieves 67\% on Pan Default and 34\% on Pan Hard,
slightly below Corrective (72\% and 36\%). These differences are modest, and the key
takeaway is that \textsc{VR-DAgger} approaches the quality of full human oversight
while reducing per-sample collection time by roughly 40\%
(Sec.~\ref{subsec:collection_efficiency}), making it the more practical choice when
supervisor time is the binding constraint.

\begin{figure}[b!]
    \vspace{-7mm}
    \centering
    \includegraphics[width=1.0\linewidth]{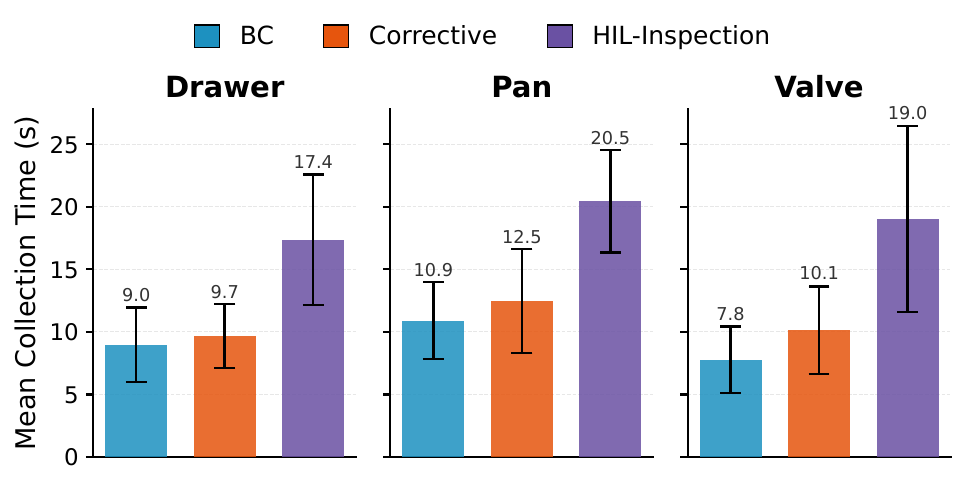}
    \caption{\textbf{Average Data Collection Effort.} Mean collection time per sample (s/trajectory) across three tasks (Pan, Drawer, Valve) for Corrective (\textsc{VR-DAgger}), HIL, and Behavioral Cloning (BC). Corrective consistently requires less collection time than HIL across all tasks, as supervisors review only short uncertainty-selected snippets rather than full rollouts, while remaining comparable to or slightly above Behavioral Cloning collection times.}
    \label{fig:data_timings}
    \vspace{-5mm}
\end{figure}
% \noindent\textbf{Default configurations show consistent gains.}
% On Default difficulty, Behavioral Cloning already achieves moderate to high success:
% 87\% (Drawer) and 49\% (Pan) with 50 demonstrations. Corrective further
% improves both: Drawer Default reaches 97\% and Pan Default reaches 54\% with
% 50 corrective segments. Task completion times remain stable across all conditions
% (Table~\ref{tab:vrdagger_results}), confirming that corrective relabeling improves
% robustness without sacrificing execution efficiency.

% \noindent\textbf{Subtask analysis reveals where failures occur.}
% Fig.~\ref{fig:performance} also reports subtask success rates alongside final task
% success rates. For Pan, the subtask success rate (reaching and grasping the pan)
% substantially exceeds the final task success rate (placing it on the burner) across all
% methods, indicating that policy failures concentrate in the later placement phase.
% Corrective relabeling closes this gap more than either HIL or scaled Behavioral Cloning, suggesting
% that uncertainty-guided snippets effectively surface the critical placement failure modes
% for targeted human correction.
% \todo{Discuss performance difference for valve task}
% \todo{Discuss subtask performance}

\noindent\textbf{Subtask analysis and task difficulty.}
Subtask success rates (Fig.~\ref{fig:performance}, Table~\ref{tab:vrdagger_results})
reveal where failures concentrate. For Pan, subtask SR (grasp success) consistently
exceeds final task SR by a large margin, e.g., BC at 150 achieves 70\% grasp but only 50\%
full-task success on the Default Pan task, indicating that failures cluster in the later placement
phase rather than the approach. Corrective relabeling narrows this gap more than scaling
BC, suggesting that uncertainty-guided snippets preferentially surface placement failures
for targeted correction. For Drawer and Valve, subtask and task SR track closely,
meaning that initiating contact or rotation is the primary bottleneck; once that is
achieved, task completion follows reliably. This explains the large Hard-difficulty gains
on Valve (+38\,pp (corrective), 44\,pp (HIL)): as the demos address a single, well-defined failure mode and
generalize effectively. Pan, by contrast, benefits from more budget: with 100 corrective
segments Pan Hard reaches 36\%, but further improvement will likely require dedicated
late-phase corrections.

\noindent\textbf{Comparison at equal total sample budget.}
Fig.~\ref{fig:performance} directly compares BC\,150, HIL\,(50\,+\,100), and
Corrective\,(50\,+\,100), all using 150 total demonstrations. Both on-policy methods
consistently outperform BC on Hard configurations by 6--14\,pp, confirming that on-policy
coverage of the failure regime cannot be replicated by simply collecting more offline
data. HIL has a slight edge on Drawer Hard (92\% vs.\ 86\%) and Valve Hard (98\%
vs.\ 92\%), where full rollout visibility helps cover more failure modes, while
Corrective leads on Pan Hard (36\% vs.\ 34\%). On Default difficulty, Drawer and Valve performance is largely
saturated across all three methods; Pan remains the exception, where on-policy collection
still yields a meaningful improvement (+22\,pp for Corrective over BC\,150).

% \noindent\textbf{Pan is the most challenging task.}
% Pan consistently yields the lowest success rates across all methods and difficulty
% levels, owing to its long-horizon structure requiring precise grasping followed by
% accurate placement. With 50 corrective segments (100 total), Pan Hard reaches 20\%; doubling the
% corrective budget to 100 segments (150 total) raises this to 40\%, indicating that this
% task benefits from a larger intervention budget, longer prediction horizons, or dedicated
% late-phase corrections. We observe greater sensitivity
% to grasp-pose variation and object clutter, which likely amplifies compounding errors
% over the long horizon.

\begin{figure}[b!]
    \vspace{-7mm}
    \hspace{-6mm}
    \centering
    \includegraphics[width=1.04\linewidth]{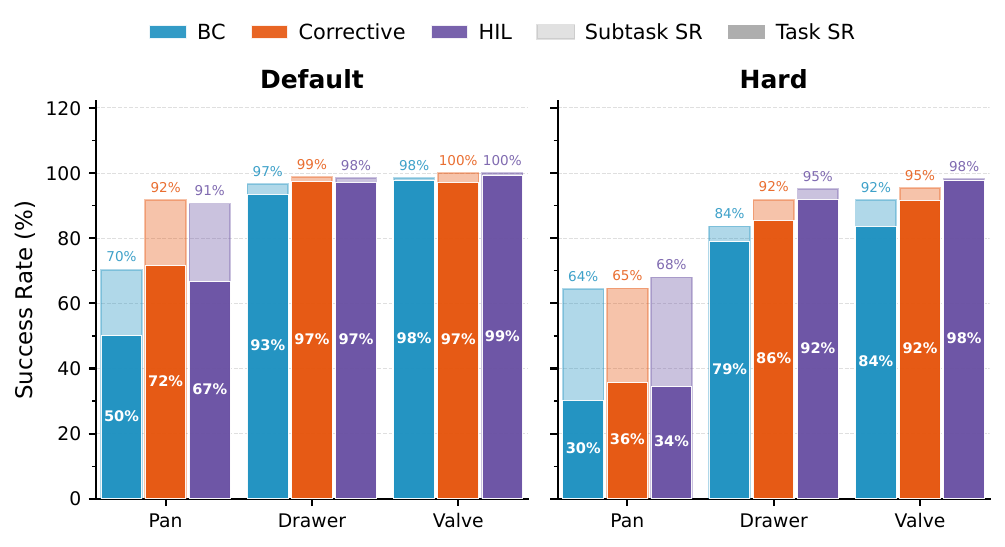}
    \caption{\textbf{Task and Subtask Success Rates.} Darker bars show final task
    success rate; lighter bars show subtask success rate (grasp success for Pan,
    contact success for Drawer and Valve), both evaluated at the end of training (1200 epochs). Results for
    \emph{Behavioral Cloning} (BC), \emph{Corrective}, and \emph{HIL} on Medium
    and Hard difficulty. Error bars show variability across evaluation rollouts.
    }
    \vspace{-5mm}
    
    \label{fig:performance}
\end{figure}

\subsection{Data Collection Efficiency}\label{subsec:collection_efficiency}

Fig.~\ref{fig:data_timings} shows mean data collection time per trajctory across all three
tasks for Corrective, HIL, and Behavioral Cloning. Corrective and BC requires significantly less
time per trajectory than HIL across all tasks (12.5 vs.\ 20.5\,s on Pan, 9.7 vs.\
17.4\,s on Drawer, 10.1 vs.\ 19.0\,s on Valve), a reduction of roughly
40\% on average. This efficiency gain stems directly from the active relabeling
design: rather than watching full rollouts to identify failure points, the supervisor
reviews only 2-second uncertainty-selected snippets before deciding where to intervene.

Corrective collection times are slightly higher than Behavioral Cloning across all tasks, which is
expected since active relabeling involves reviewing snippets and taking over from a
mid-episode state, whereas Behavioral Cloning demonstrations always start from a clean initial
configuration. The overhead is modest (0.7--2.3\,s/sample above Behavioral Cloning, per
Fig.~\ref{fig:data_timings}) and is largely compensated by the performance
improvements reported in Table~\ref{tab:vrdagger_results}.

\section{Conclusion}\label{sec:conclusion}

We present \textsc{VR-DAgger}, a human-in-the-loop imitation learning framework that
combines immersive VR teleoperation with uncertainty-guided on-policy data collection.
The framework is built around a modular, task-agnostic VR application that integrates
directly with Isaac Lab. Capable of collecting data with any USD scene with minimal overhead from a training server, allowing easy integration with existing learning stacks.
Rather than requiring continuous supervision, the system decouples policy
execution from human attention: the policy rolls out autonomously in simulation, and MC
dropout uncertainty surfaces the most informative failure segments as short VR snippets
for the operator to review and correct, concentrating effort precisely where the policy
is most uncertain.

Evaluated on three dexterous manipulation tasks (Pan, Drawer, Valve) with the 10-DoF
XHand, active corrective labeling consistently improves over Behavioral Cloning across
all tasks and difficulty levels. At matched total budget (150 samples), both on-policy
methods outperform BC\,150 on Hard configurations by 6--14\,pp; on Default difficulty,
Drawer and Valve saturate across all methods while Pan remains the exception, with
Corrective yielding a substantial gain of +22\,pp. This highlights the sample efficiency
of on-policy correction particularly where performance is not yet saturated. Beyond
accuracy, \textsc{VR-DAgger} reduces per-sample collection time by approximately 40\%
compared to unguided HIL, as operators review short uncertainty-selected snippets rather
than full rollouts, making it the more practical choice when supervisor time is the
binding constraint.
% Evaluated on three dexterous manipulation tasks (Pan, Drawer, Valve) using the
% 10-DoF XHand under standard and challenging initial configurations, corrective labeling
% consistently improves over Behavioral Cloning across all tasks and difficulty levels.
% The largest gains arise on challenging configurations: Valve Hard improves from 54\% to
% 92\% with 50 corrective segments (+38\,pp), and  corrective labeling with the same total
% budget as Behavioral Cloning (Corrective 50\,+\,100 vs.\ BC\,150) outperforms scaled offline data by
% up to 22\,pp. At the same time, corrective labeling reduces per-sample collection cost by
% approximately 40\% compared to Human-in-the-loop labeling, since operators review targeted snippets rather than full rollouts. 
% Together, these results demonstrate that uncertainty-guided correction
% offers a more data-efficient path to robust visuomotor policies than scaling offline
% demonstrations alone.

Our evaluation is currently limited to simulation and a single dexterous hand embodiment.
Future work will pursue two directions. First, we will use the framework to deploy policies on physical
hardware, transferring learned policies to real robots and closing the sim-to-real gap.
Second, since all execution runs in simulation, we plan to exploit this by applying
aggressive domain randomization over textures, lighting, and object appearance to improve
generalization at low additional cost.
\vspace{-5mm}

%%%%%%%%%%%%%%%%%%%%%%%%%%%%%%%%%%%%%%%%%%%%%%%%%%%%%%%%%%%%%%%%%%%%%%%%%%%%%%%%
% \input{chapters/A_appendix}

\endgroup

%%%%%%%%%%%%%%%%%%%%%%%%%%%%%%%%%%%%%%%%%%%%%%%%%%%%%%%%%%%%%%%%%%%%%%%%%%%%%%%%

% BibTeX needs a bibliography style; IEEEconf templates typically use IEEEtran.bst
\bibliographystyle{IEEEtran}
\bibliography{bibtex}

\end{document}